\documentclass{article}

\usepackage[accepted]{icml2010}
\usepackage{graphicx}
\usepackage{amsmath}
\usepackage{amssymb}
\usepackage{amsthm}
\usepackage{psfrag}
\usepackage{booktabs}
\usepackage{natbib}
\newcommand{\field}[1]{\mathbb{#1}}

\newcommand{\fs}[1]{\mathcal{#1}}

\newcommand{\nats}{\field{N}}

\newcommand{\define}{:=}

\newcommand{\prob}{\mathbf{P\!r}}      % Probability
\newcommand{\g}[1]{\underline{#1}}

\theoremstyle{plain}
\newtheorem{theorem}{Theorem}

\theoremstyle{definition}

\icmltitlerunning{Convergence of Bayesian Control Rule}

\begin{document}

%\bibliographystyle{plainnat}

%\title{A Bayesian Rule for Adaptive Control based on Causal Interventions}

\comment{
\author{Pedro A. Ortega\\
        Department of Engineering\\
        University of Cambridge\\
        Cambridge CB2 1PZ, UK\\
        \texttt{peortega@dcc.uchile.cl}\\
        \And
        Daniel A. Braun\\
        Department of Engineering\\
        University of Cambridge\\
        Cambridge CB2 1PZ, UK\\
        \texttt{dab54@cam.ac.uk}\\
} }

\twocolumn[ \icmltitle{Convergence of the Bayesian Control Rule}

\icmlauthor{Pedro A. Ortega}{peortega@dcc.uchile.cl} % LEAVE BLANK FOR ORIGINAL SUBMISSION
\icmlauthor{Daniel A. Braun}{dab54@cam.ac.uk} % LEAVE BLANK FOR ORIGINAL SUBMISSION
\icmladdress{Dept. of Engineering,
    University of Cambridge,
    Cambridge CB2~1PZ, UK}
\vskip 0.3in ]

%\author{Anonymous}

%\makeanontitle
%\maketitle

\begin{abstract}%
Recently, new approaches to adaptive control have sought to reformulate the
problem as a minimization of a relative entropy criterion to obtain tractable
solutions. In particular, it has been shown that minimizing the expected
deviation from the causal input-output dependencies of the true plant leads to
a new promising stochastic control rule called the Bayesian control rule. This
work proves the convergence of the Bayesian control rule under two sufficient
assumptions: boundedness, which is an ergodicity condition; and consistency,
which is an instantiation of the sure-thing principle.
\end{abstract}

\emph{Keywords:} Adaptive behavior, Intervention calculus, Bayesian control, Kullback-Leibler-divergence

\section{Introduction}

When the behavior of a plant under any control signal is fully known, then the
designer can choose a controller that produces the desired dynamics. Instances
of this problem include hitting a target with a cannon under known weather
conditions, solving a maze having its map and controlling a robotic arm in a
manufacturing plant. However, when the behavior of the plant is unknown, then
the designer faces the problem of adaptive control. For example, shooting the
cannon lacking the appropriate measurement equipment, finding the way out of an
unknown maze and designing an autonomous robot for Martian exploration.
Adaptive control turns out to be far more difficult than its non-adaptive
counterpart. Even when the plant dynamics is known to belong to a particular
class for which optimal controllers are available, constructing the
corresponding optimal adaptive controller is in general intractable even for
simple toy problems \citep{Duff2002}. Thus, virtually all of the effort of the
research community is centered around the development of tractable
approximations.

Recently, new formulations of the adaptive control problem that are based on
the minimization of a relative entropy criterion have attracted the interest of
the control and reinforcement learning community. For example, it has been
shown that a large class of optimal control problems can be solved very
efficiently if the problem statement is reformulated as the minimization of the
deviation of the dynamics of a controlled system from the uncontrolled system
\citep{Todorov2006, Todorov2009, Kappen2009}. A similar approach minimizes the
deviation of the causal input/output-relationship of a Bayesian mixture of
controllers from the true controller, obtaining an explicit solution called the
\emph{Bayesian control rule} \citep{OrtegaBraun2010}. This control rule is
particularly interesting because it leads to stochastic controllers that infer
the optimal controller on-line by combining the plant-specific controllers,
implicitly using the uncertainty of the dynamics to trade-off exploration
versus exploitation.

Although the Bayesian control rule constitutes a promising approach to adaptive
control, there are currently no proofs that guarantee its convergence to the
desired policy. The aim of this paper is to develop a set of sufficient
conditions of convergence and then to provide a proof. The analysis is limited
to the simple case of controllers having a finite amount of modes of operation.
Special care has been taken to illustrate the motivation behind the concepts.

\section{Preliminaries}

The exposition is restricted to the case of discrete time with discrete
stochastic observations and control signals. Let $\fs{O}$ and $\fs{A}$ be two
finite sets of symbols, where the former is the set of inputs (observations)
and the second the set of outputs (actions). Actions and observations at time
$t$ are denoted as $a_t \in \fs{A}$ and $o_t \in \fs{O}$ respectively, and the
shorthand $a_{\leq t} \define a_1, a_2, \ldots, a_t$ and the like are used to
simplify the notation of strings. Symbols are underlined to glue them together
as in $\g{ao}_{\leq t} = a_1, o_1, a_2, o_2, \ldots, a_t, o_t$. It is assumed
that the interaction between the controller and the plant proceeds in cycles
$t=1, 2, \ldots$ where in cycle $t$ the controller issues action $a_t$ and the
plant responds with an observation $o_t$.

A \emph{controller} is defined as a probability distribution $P$ over the
input/output (I/O) stream, and it is fully characterized by the conditional
probabilities
\[
    P(a_t|\g{ao}_{<t})
    \quad\text{and}\quad
    P(o_t|\g{ao}_{<t}a_t)
\]
representing the probabilities of emitting action $a_t$ and collecting
observation $o_t$ given the respective I/O history. Similarly, a \emph{plant}
is defined as a probability distribution $Q$ characterized by the conditional
probabilities
\[
    Q(o_t|\g{ao}_{<t}a_t)
\]
representing the probabilities of emitting observation $o_t$ given the I/O
history.

If the plant is known, i.e. if the conditional probabilities
$Q(o_t|\g{ao}_{<t}a_t)$ are known, then the designer can build a suitable
controller by equating the observation streams as $P(o_t|\g{ao}_{<t}a_t) =
Q(o_t|\g{ao}_{<t}a_t)$ and by defining action probabilities
$P(a_t|\g{ao}_{<t})$ such that the resulting distribution $P$ maximizes a
desired utility criterion. In this case $P$ is said to be \emph{tailored to}
$Q$. In many situations the conditional probabilities $P(a_t|\g{ao}_{<t})$ will
be deterministic, but there are cases (e.g. in repeated games) where the
designer might prefer stochastic policies instead.

If the plant is unknown then one faces an adaptive control problem. Assume we
know that the plant $Q_m$ is going to be drawn randomly from a set $\fs{Q}
\define \{ Q_m \}_{m \in \fs{M}}$ of possible plants indexed by $\fs{M}$.
Assume further we have available a set of controllers $\fs{P}
\define \{ P_m \}_{m \in \fs{M}}$, where each $P_m$ is tailored to $Q_m$.
How can we now construct a controller $P$ such that its behavior is as close as
possible to the tailored controller $P_m$ under any realization of $Q_m \in
\fs{Q}$?

\section{Bayesian Control Rule}

A na\"{\i}ve approach would be to minimize the relative entropy of the
controller $P$ with respect to the true controller $P_m$, averaged over all
possible values of $m$. However, this is syntactically incorrect. The important
observation made in \citet{OrtegaBraun2010} is that we do not want to minimize
the deviation of $P$ from $P_m$, but the deviation of the causal I/O
dependencies in $P$ from the causal I/O dependencies in $P_m$. Intuitively
speaking, one does not want to predict actions and observations, but to predict
the observations (effect) given actions (causes). More specifically, they
propose to minimize a set of (causal) divergences $C$ defined by
\begin{equation}\label{eq:causal-divergence}
    C \define \limsup_{t \rightarrow \infty}
        \sum_m P(m) \sum_{\tau=1}^t C_\tau \\
\end{equation}
where
\begin{align*}
    C_\tau &\define
    \sum_{o_{<\tau}} P_m(\g{\hat{a}o}_{<\tau})
    C_\tau(\g{\hat{a}o}_{<\tau}) \\
    C_\tau(h) & \define
        \sum_{a_\tau} \sum_{o_\tau} P_m(\g{ao}_\tau|h)
        \log \frac{ P_m(\g{ao}_\tau|h) }
                  { P(\g{ao}_\tau|h) },
\end{align*}
and where $P(m)$ is the prior probability of $m \in \fs{M}$, $\hat{a}_\tau$
denotes an intervened (not observed) action at time $\tau$, and $\hat{a}_1,
\hat{a}_2, \hat{a}_3, \ldots$ is an arbitrary sequence of intervened actions
that gives rise to a particular instantiation of $C$.

In \citet{OrtegaBraun2010}, it is shown that the controller $P$ that minimizes
$C$ in Equation~(\ref{eq:causal-divergence}) for any sequence of intervened
actions is given by the conditional probabilities
\begin{equation}\label{eq:bcr-conditionals}
\begin{aligned}
    P(a_t|\g{\hat{a}o}_{<\tau}) &\define
    \sum_m P_m(a_t|\g{ao}_{<\tau}) P(m|\g{\hat{a}o}_{<\tau}) \\
    P(o_t|\g{\hat{a}o}_{<\tau}) &\define
    \sum_m P_m(o_t|\g{ao}_{<\tau}a_\tau) P(m|\g{\hat{a}o}_{<\tau}) \\
\end{aligned}
\end{equation}
where
\begin{equation}\label{eq:bcr-posterior}
    P(m|\g{\hat{a}o}_{\leq t}) \define
        \frac{ P_m(o_t|\g{ao}_{<t}a_t) P(m|\g{\hat{a}o}_{<t}) }
             { \sum_{m'} P_{m'}(o_t|\g{ao}_{<t}a_t) P(m'|\g{ao}_{<t}) }.
\end{equation}
Equations~(\ref{eq:bcr-conditionals}) and~(\ref{eq:bcr-posterior}) constitute
the \emph{Bayesian control rule}. This result is obtained by using properties
of interventions using causal calculus \citep{Pearl2000}. It is worth to point
out that the resulting controller is fully defined in terms of its constituent
controllers in $\fs{P}$. It is customary to use the notation
\begin{align*}
    P(a_t|m, \g{ao}_{<t}) &\define P_m(a_t|\g{ao}_{<t})
    \\
    P(o_t|m, \g{ao}_{<t}a_t) &\define P_m(o_t|\g{ao}_{<t}a_t),
\end{align*}
that is, treating the different controllers as ``hypotheses'' of a Bayesian
model. In the context of the Bayesian control rule, these ``I/O hypotheses''
are called \emph{operation modes}. Note that the resulting control law is in
general stochastic.

\section{Policy Diagrams}

\begin{figure}[htbp]
\centering %
\begin{scriptsize}
\psfrag{sp}[c]{state space} %
\psfrag{po}[c]{policy} %
\psfrag{s1}[c]{$s$} %
\psfrag{s2}[c]{$s'$} %
\psfrag{st}[c]{$ao$} %
\includegraphics[width=7cm]{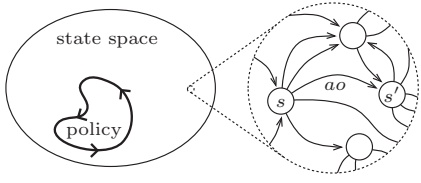}
\end{scriptsize}
\caption{A policy diagram.}
\label{fig:policy-diagram} %
\end{figure}

A policy diagram is a useful informal tool to analyze the effect of control
policies on plants. Figure~\ref{fig:policy-diagram}, illustrates an example.
One can imagine a plant as a collection of states connected by transitions
labeled by I/O symbols. For instance, Figure~\ref{fig:policy-diagram}
highlights a state $s$ where taking action $a \in \fs{A}$ and collecting
observation $o \in \fs{O}$ leads to state $s'$. In a policy diagram, one
abstracts away from the underlying details of the plant's dynamics,
representing sets of states and transitions as enclosed areas similar to a Venn
diagram. Choosing a particular policy in a plant amounts to partially
controlling the transitions taken in the state space, thereby choosing a subset
of the plant's dynamics. Accordingly, a policy is represented by a subset in
state space (enclosed by a directed curve) as illustrated in
Figure~\ref{fig:policy-diagram}.

Policy diagrams are especially useful to analyze the effect of policies on
different hypotheses about the plant's dynamics. A controller that is endowed
with a set of operation modes $\fs{M}$ can be seen as having \emph{hypotheses}
about the plant's underlying dynamics, given by the observation models
$P(o_t|m,\g{ao}_{<t}a_t)$, and associated \emph{policies}, given by the action
models $P(a_t|m,\g{ao}_{<t})$, for all $m \in \fs{M}$. For the sake of
simplifying the interpretation of policy diagrams, we will assume\footnote{Note
however that no such assumptions are made to obtain the results of this paper.}
the existence of a state space $\fs{S}$ and a function $T: (\fs{A}\times\fs{O})
\rightarrow \fs{S}$ mapping I/O histories into states. With this assumption,
policies and hypotheses can be seen as conditional probabilities
\begin{align*}
    P(a_t|m,s) &\define P(a_t|m,\g{ao}_{<t})
    \\ \text{and }P(o_t|m,s,a_t) &\define P(o_t|m,\g{ao}_{<t}a_t)
\end{align*}
respectively, defining transition probabilities
\[
    P(s'|m,s) = \sum_{\fs{S'}} P(\g{ao}_t|m,s)
\]
for a Markov chain in the state space, where $s = T(\g{ao}_{<t})$ and $\fs{S}'$
contains the transitions $\g{ao}_t$ such that $T(\g{ao}_{\leq t}) = s'$.

\section{Divergence Processes}

One of the obvious questions to ask oneself with respect to the Bayesian
control rule is whether it converges to the right control law or not. That is,
whether $P(a_t|\g{\hat{a}o}_t) \rightarrow P(a_t|m^\ast,\g{ao}_{<t})$ as $t
\rightarrow \infty$ when $m^\ast$ is the true operation mode, i.e. the
operation mode such that $P(a_t|m^\ast,\g{ao}_{<t}) = Q(a_t|\g{ao}_{<t})$. As
will be obvious from the discussion in the rest of this paper, this is in
general not true.

As it is easily seen from Equation~\ref{eq:bcr-conditionals}, showing
convergence amounts to show that the posterior distribution
$P(m|\g{\hat{a}o}_{<t})$ concentrates its probability mass on a subset of
operation modes $\fs{M}^\ast$ having essentially the same output stream as
$m^\ast$,
\begin{align*}
    \sum_{m \in \fs{M}}
        & P(a_t|m,\g{ao}_{<t}) P(m|\g{\hat{a}o}_{<t})
    \\ & \approx \sum_{m \in \fs{M}^\ast}
        P(a_t|m^\ast,\g{ao}_{<t}) P(m|\g{\hat{a}o}_{<t})
    \\ & \approx
        P(a_t|m^\ast,\g{ao}_{<t}).
\end{align*}

\begin{figure}[htbp]
\centering %
\begin{scriptsize}
\psfrag{or}{0} %
\psfrag{ca}{1} %
\psfrag{cb}{2} %
\psfrag{cc}{3} %
\psfrag{cd}{4} %
\psfrag{lx}{$t$} %
\psfrag{ly}{$d_t$} %
\includegraphics[width=7cm]{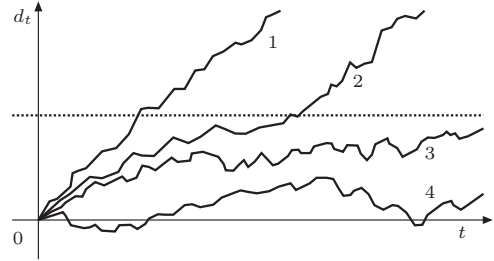}
\end{scriptsize}
\caption{Realization of the divergence processes~1 to~4 associated to a
controller with operation modes $m_1$ to $m_4$. The divergence processes~1
and~2 diverge, whereas~3 and~4 stay below the dotted bound. Hence, the
posterior probabilities of $m_1$ and $m_2$ vanish.}
\label{fig:divergences} %
\end{figure}

Hence, understanding the asymptotic behavior of the posterior probabilities
\[
    P(m|\g{\hat{a}o}_{\leq t})
\]
is the main goal of this paper. In particular, one wants to understand under
what conditions these quantities converge to zero. The posterior can be
rewritten as
\begin{align*}
    P(m|\g{\hat{a}o}_{\leq t})
    &= \frac{ P(\g{\hat{a}o}_{\leq t}|m) P(m) }
           { \sum_{m' \in \fs{M}} P(\g{\hat{a}o}_{\leq t}|m') P(m') }
    \\&= \frac{ P(m) \prod_{\tau=1}^t P(o_\tau|m,\g{ao}_{<\tau}a_\tau) }
           { \sum_{m' \in \fs{M}}
             P(m') \prod_{\tau=1}^t P(o_\tau|m',\g{ao}_{<\tau}a_\tau) }.
\end{align*}
If all the summands but the one with index $m^\ast$ are dropped from the
denominator, one obtains the bound
\begin{align*}
    P(m|\g{\hat{a}o}_{\leq t})
    \leq \ln \frac{ P(m) }{ P(m^\ast) } \prod_{\tau=1}^t
        \frac{ P(o_\tau|\g{ao}_{<\tau}a_\tau|m) }
             { P(o_\tau|\g{ao}_{<\tau}a_\tau|m^\ast) },
\end{align*}
which is valid for all $m^\ast \in \fs{M}$. From this inequality, it is seen
that it is convenient to analyze the behavior of the stochastic process
\[
    d_t(m^\ast\|m) \define \sum_{\tau=1}^t
        \ln \frac{ P(o_\tau|m^\ast, \g{ao}_{<\tau}a_\tau) }
                 { P(o_\tau|m, \g{ao}_{<\tau}a_\tau) }
\]
which is the \emph{divergence process} of $m$ from the reference $m^\ast$.
Indeed, if $d_t(m^\ast\|m) \rightarrow \infty$ as $t \rightarrow \infty$, then
\begin{align*}
    \lim_{t \rightarrow \infty} &
    \frac{ P(m) }{ P(m^\ast) } \prod_{\tau=1}^t
        \frac{ P(o_\tau|\g{ao}_{<\tau}a_\tau|m) }
             { P(o_\tau|\g{ao}_{<\tau}a_\tau|m^\ast) }
    \\&= \lim_{t \rightarrow \infty}
        \frac{P(m)}{P(m^\ast)} \cdot e^{-d_t(m^\ast\|m)}
    = 0,
\end{align*}
and thus clearly $P(m|\g{\hat{a}o}_{\leq t}) \rightarrow 0$.
Figure~\ref{fig:divergences} illustrates simultaneous realizations of the
divergence processes of a controller. Intuitively speaking, these processes
provide lower bounds on accumulators of surprise value measured in information
units.

\begin{figure}[htbp]
\centering %
\begin{scriptsize}
\psfrag{or}{0} %
\psfrag{ca}{1} \psfrag{pd}{1} %
\psfrag{cb}{2} \psfrag{pb}{2} %
\psfrag{cc}{3} \psfrag{pc}{3} %
\psfrag{lx}{$t$} %
\psfrag{ly}{$d_t$} %
\includegraphics[width=7cm]{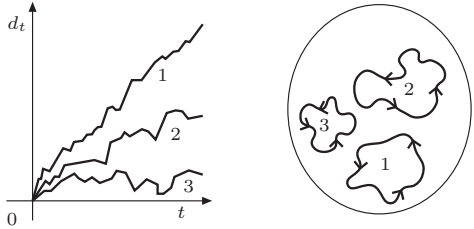}
\end{scriptsize}
\caption{The application of different policies lead to different statistical
properties of the same divergence process.}
\label{fig:policy2divergence} %
\end{figure}

A divergence process is a random walk, i.e. whose value at time $t$ depends on
the whole history up to time $t-1$. What makes them cumbersome to characterize
is the fact that their statistical properties depend on the particular policy
that is applied; hence, a given divergence process can have different growth
rates depending on the policy (Figure~\ref{fig:policy2divergence}). Indeed, the
behavior of a divergence process might depend critically on the distribution
over actions that is used. For example, it can happen that a divergence process
stays stable under one policy, but diverges under another. In the context of
the Bayesian control rule this problem is further aggravated, because in each
time step, the policy to apply is determined stochastically. More specifically,
if $m^\ast$ is the true operation mode, then $d_t(m^\ast\|m)$ is a random
variable that depends on the realization $\g{ao}_{\leq t}$ which is drawn from
\begin{align*}
    \prod_{\tau=1}^t
        P(a_\tau|m_\tau,\g{ao}_{\leq \tau})
        P(o_\tau|m^\ast,\g{ao}_{\leq \tau}a_\tau),
\end{align*}
where the $m_1, m_2, \ldots, m_t$ are drawn themselves from $P(m_1),
P(m_2|\g{\hat{a}o}_1), \ldots, P(m_t|\g{\hat{a}o}_{<t})$.

\begin{figure}[htbp]
\centering %
\begin{scriptsize}
\psfrag{or}{0} %
\psfrag{ca}{1} %
\psfrag{cb}{2} %
\psfrag{cc}{3} %
\psfrag{lx}{$t$} %
\psfrag{ly}{$d_t$} %
\includegraphics[width=7cm]{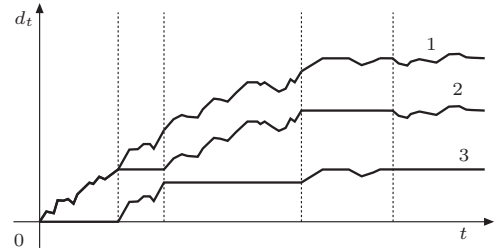}
\end{scriptsize}
\caption{Decomposition of a divergence process (1) into sub-divergences (2 \&
3).}
\label{fig:divergence-decomposition} %
\end{figure}

To deal with the heterogeneous nature of divergence processes, one can
introduce a temporal decomposition that demultiplexes the original process into
many sub-processes belonging to unique policies. Let $\fs{N}_{t} \define
\{1,2,\ldots,t\}$ be the set of time steps up to time $t$. Let $\fs{T} \subset
\fs{N}_t$, and let $m, m' \in \fs{M}$. Define a \emph{sub-divergence} of
$d_t(m\|m)$ as a random variable
\[
    g(m';\fs{T}) \define \sum_{\tau \in \fs{T}}
        \ln \frac{ P(o_\tau|m^\ast, \g{ao}_{<\tau}a_\tau) }
                 { P(o_\tau|m, \g{ao}_{<\tau}a_\tau) }
\]
drawn from
\begin{multline*}
    P^m_{m'}(\{\g{ao}_\tau\}_{\tau \in \fs{T}}|
        \{\g{ao}_\tau\}_{\tau \in \fs{T}^\complement})
    \\ \define
    \Bigl( \prod_{\tau \in \fs{T}} P(a_\tau|m, \g{ao}_{<\tau}) \Bigr)
    \Bigl( \prod_{\tau \in \fs{T}} P(o_\tau|m', \g{ao}_{<\tau}a_\tau) \Bigr),
\end{multline*}
where $\fs{T}^\complement \define \fs{N}_t \setminus \fs{T}$ and where
$\{\g{ao}_\tau\}_{\tau \in \fs{T}^\complement}$ are given conditions that are
kept constant. In this definition, $m'$ plays the role of the policy that is
used to sample the actions in the time steps $\fs{T}$. Clearly, any realization
of the divergence process $d_t(m^\ast\|m)$ can be decomposed into a sum of
sub-divergences, i.e.
\begin{equation}\label{eq:divergence-decomposition}
    d_t(m^\ast\|m)
    = \sum_{m'} g(m';\fs{T}_{m'}),
\end{equation}
where $\{ \fs{T}_{m} \}_{m \in \fs{M}}$ forms a partition of $\fs{N}_t$.
Figure~\ref{fig:divergence-decomposition} shows an example decomposition.

The averages of sub-divergences will play an important r\^{o}le in the
analysis. Define the average over all realizations of $g(m';\fs{T})$ as
\begin{multline*}
    G(m',\fs{T})
    \\ \define
    \sum_{(\g{ao}_\tau)_{\tau \in \fs{T}}}
    P^m_{m'}(\{\g{ao}_\tau\}_{\tau \in \fs{T}}|
        \{\g{ao}_\tau\}_{\tau \in \fs{T}^\complement})
    g(m';\fs{T}).
\end{multline*}
Notice that for any $\tau \in \fs{N}_t$,
\begin{multline*}
    G(m';\{\tau\})
    \\= \sum_{\g{ao}_\tau}
        P(a_\tau|m',\g{ao}_{<\tau})
        P(o_\tau|m^\ast,\g{ao}_{<\tau}a_\tau)
    \\ \cdot \ln
        \frac{ P(o_\tau|m^\ast,\g{ao}_{<\tau}a_\tau) }
             { P(o_\tau|m,\g{ao}_{<\tau}a_\tau) }
    \geq 0,
\end{multline*}
because of Gibbs' inequality. In particular,
\[
    G(m^\ast;\{\tau\}) = 0.
\]
Clearly, this holds as well for any $\fs{T} \subset \fs{N}_t$:
\begin{equation}
\begin{aligned}\label{eq:gibbs-ineq-sub-diver}
    \forall m' \quad G(m';\fs{T}) & \geq 0,\\
    G(m^\ast;\fs{T}) & = 0.
\end{aligned}
\end{equation}

\section{Boundedness}

In general, a divergence process is very complex: virtually all the classes of
distributions that are of interest in control go well beyond \mbox{i.i.d.} and
stationary processes. This increased complexity can jeopardize the analytic
tractability of the divergence process, i.e. such that no predictions about its
asymptotic behavior can be made anymore. More specifically, if the growth rates
of the divergence processes vary too much from realization to realization, then
the posterior distribution over operation modes can vary qualitatively between
realizations. Hence, one needs to impose a stability requirement akin to
ergodicity to limit the class of possible divergence-processes to a class that
is analytically tractable. In the light of this insight, the following property
is introduced.

A divergence process $d_t(m^\ast\|m)$ is said to be \emph{bounded} in $\fs{M}$
iff for any $\delta > 0$, there is a $C \geq 0$, such that for all $m' \in
\fs{M}$, all $t$ and all $\fs{T} \subset \fs{N}_t$
\[
    \Bigl|
        g(m';\fs{T}) - G(m';\fs{T})
    \Bigr|
    \leq C
\]
with probability $\geq 1-\delta$.

\begin{figure}[htbp]
\centering %
\begin{scriptsize}
\psfrag{or}{0} %
\psfrag{ca}{1} %
\psfrag{cb}{2} %
\psfrag{cc}{3} %
\psfrag{lx}{$t$} %
\psfrag{ly}{$d_t$} %
\includegraphics[width=7cm]{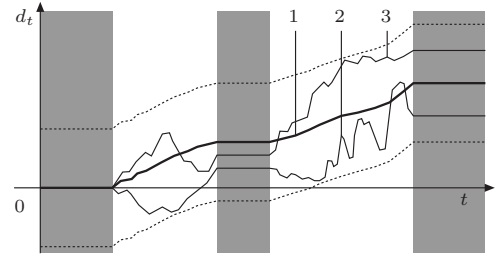}
\end{scriptsize}
\caption{If a divergence process is bounded, then the realizations (curves 2 \&
3) of a sub-divergence stay within a band around the mean (curve 1).}
\label{fig:divergence-boundedness} %
\end{figure}

Figure~\ref{fig:divergence-boundedness} illustrates this property. Boundedness
is the key property that is going to be used to construct the results of this
paper. The first important result is that the posterior probability of the true
operation mode is bounded from below.

\begin{theorem}\label{theo:lower-bound}
Let the set of operation modes of a controller be such that for all $m \in
\fs{M}$ the divergence process $d_t(m^\ast\|m)$ is bounded. Then, for any
$\delta
> 0$, there is a $\lambda > 0$, such that for all $t \in \nats$,
\[
    P(m^\ast|\g{\hat{a}o}_{\leq t}) \geq \frac{\lambda}{|\fs{M}|}
\]
with probability $\geq 1 - \delta$.
\end{theorem}
\begin{proof}
As has been pointed out in~(\ref{eq:divergence-decomposition}), a particular
realization of the divergence process $d_t(m^\ast\|m)$ can be decomposed as
\[
    d_t(m^\ast\|m)
    = \sum_{m'} g_m(m';\fs{T}_{m'}),
\]
where the $g_m(m';\fs{T}_{m'})$ are sub-divergences of $d_t(m^\ast\|m)$ and the
$\fs{T}_{m'}$ form a partition of $\fs{N}_t$. However, since $d_t(m^\ast\|m)$
is bounded in $\fs{M}$, one has for all $\delta'>0$, there is a $C(m) \geq 0$,
such that for all $m' \in \fs{M}$, all $t \in \fs{N}_t$ and all $\fs{T} \subset
\fs{N}_t$, the inequality
\[
    \Bigl|
        g_m(m';\fs{T}_{m'}) - G_m(m';\fs{T}_{m'})
    \Bigr|
    \leq C(m)
\]
holds with probability $\geq 1-\delta'$. However, due
to~(\ref{eq:gibbs-ineq-sub-diver}),
\[
    G_m(m';\fs{T}_{m'}) \geq 0
\]
for all $m' \in \fs{M}$. Thus,
\[
    g_m(m';\fs{T}_{m'}) \geq -C(m).
\]
If all the previous inequalities hold simultaneously then the divergence
process can be bounded as well. That is, the inequality
\begin{equation}\label{eq:proof-1-a}
    d_t(m^\ast\|m) \geq -M C(m)
\end{equation}
holds with probability $\geq (1-\delta')^M$ where $M \define |\fs{M}|$. Choose
\[
    \beta(m) \define \max\{0, \ln \tfrac{P(m)}{P(m^\ast)}\}.
\]
Since $0 \geq \ln \tfrac{P(m)}{P(m^\ast)} - \beta(m)$, it can be added to the
right hand side of~(\ref{eq:proof-1-a}). Using the definition of
$d_t(m^\ast\|m)$, taking the exponential and rearranging the terms one obtains
\begin{multline*}
    P(m^\ast) \prod_{\tau=1}^t P(o_\tau|m^\ast,\g{ao}_{<\tau}a_\tau)
    \\ \geq e^{-\alpha(m)} P(m) \prod_{\tau=1}^t P(o_\tau|m^\ast,\g{ao}_{<\tau}a_\tau)
\end{multline*}
where $\alpha(m) \define M C(m) + \beta(m) \geq 0$. Identifying the posterior
probabilities of $m^\ast$ and $m$ by dividing both sides by the normalizing
constant yields the inequality
\[
    P(m^\ast|\g{\hat{a}o}_{\leq t}) \geq e^{-\alpha(m)} P(m|\g{\hat{a}o}_{\leq t}).
\]
This inequality holds simultaneously for all $m \in \fs{M}$ with probability
$\geq (1-\delta')^{M^2}$ and in particular for $\lambda \define \min_m
\{e^{-\alpha(m)}\}$, that is,
\[
    P(m^\ast|\g{\hat{a}o}_{\leq t}) \geq \lambda P(m|\g{\hat{a}o}_{\leq t}).
\]
But since this is valid for any $m \in \fs{M}$, and because $\max_m \{
P(m|\g{\hat{a}o}_{\leq t}) \} \geq \frac{1}{M}$, one gets
\[
    P(m^\ast|\g{\hat{a}o}_{\leq t}) \geq \frac{\lambda}{M},
\]
with probability $\geq 1-\delta$ for arbitrary $\delta > 0$ related to
$\delta'$ through the equation $\delta' \define 1 - \sqrt[M^2]{1-\delta}$.
\end{proof}

\section{Core}

\begin{figure}[htbp]
\centering %
\begin{scriptsize}
\psfrag{h1}{$H_1$} \psfrag{p1}{$P_1$} %
\psfrag{h2}{$H_2$} \psfrag{p2}{$P_2$} %
\psfrag{h3}{$H_3$} \psfrag{p3}{$P_3$} %
\psfrag{ra1}{$A$} \psfrag{ra2}{$A$} %
\psfrag{rb1}{$B$} \psfrag{rb2}{$B$} %
\includegraphics[width=7cm]{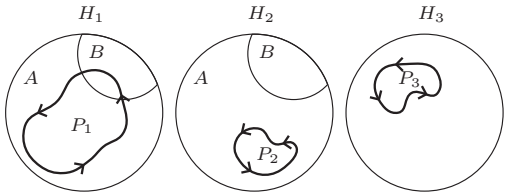}
\end{scriptsize}
\caption{If hypothesis $H_1$ is true and agrees with $H_2$ on region $A$, then
policy $P_2$ cannot disambiguate the three hypotheses.}
\label{fig:core} %
\end{figure}

If one wants to identify the operation modes whose posterior probabilities
vanish, then it is not enough to characterize them as those whose hypothesis
does not match the true hypothesis. Figure~\ref{fig:core} illustrates this
problem. Here, three hypotheses along with their associated policies are shown.
$H_1$ and $H_2$ share the prediction made for region~$A$ but differ in
region~$B$. Hypothesis $H_3$ differs everywhere from the others. Assume $H_1$
is true. As long as we apply policy~$P_2$, hypothesis~$H_3$ will make wrong
predictions and thus its divergence process will diverge as expected. However,
no evidence against~$H_2$ will be accumulated. It is only when we apply
policy~$P_1$ \emph{for long enough time} that the controller will eventually
enter region~$B$ and hence accumulate counter-evidence for $H_2$.

But what does ``long enough'' mean? If $P_1$ is executed only for a short
period, then the controller risks not visiting the disambiguating region. But
unfortunately, neither the right policy nor the right length of the period to
run it are known beforehand. Hence, the controller needs a clever
time-allocating strategy to test all policies for all finite time intervals.
This motivates following definition.

The \emph{core} of an operation mode $m^\ast$, denoted as $[m^\ast]$, is the
subset of $\fs{M}$ containing operation modes behaving like $m^\ast$ under its
policy. More formally, an operation mode $m \notin [m^\ast]$ (i.e. is
\emph{not} in the core) iff for any $C \geq 0$, $\delta, \xi
> 0$, there is a $t_0 \in \nats$, such that for all $t \geq t_0$,
\[
    G(m^\ast;\fs{T}) \geq C
\]
with probability $\geq 1-\delta$, where $G(m^\ast;\fs{T})$ is a sub-divergence
of $d_t(m^\ast\|m)$, and $\prob\{\tau \in \fs{T}\} \geq \xi$ for all $\tau \in
\fs{N}_t$.

In other words, if the controller was to apply $m^\ast$'s policy in each time
step with probability at least $\xi$, and under this strategy the expected
sub-divergence $G(m^\ast;\fs{T})$ of $d_t(m^\ast\|m)$ grows unboundedly, then
$m$ is not in the core of $m^\ast$. Note that demanding a strictly positive
probability of execution in each time step guarantees that controller will run
$m^\ast$ for all possible finite time-intervals. As the following theorem
shows, the posterior probabilities of the operation modes that are not in the
core vanish almost surely.

\begin{theorem}\label{theo:vanishing-posterior}
Let the set of operation modes of a controller be such that for all $m \in
\fs{M}$ the divergence process $d_t(m^\ast\|m)$ is bounded. Then, if $m \notin
[m^\ast]$, then $P(m|\g{\hat{a}o}_{\leq t}) \rightarrow 0$ as $t \rightarrow
\infty$ almost surely.
\end{theorem}

\begin{proof}
The divergence process $d_t(m^\ast\|m)$ can be decomposed into a sum of
sub-divergences (see Equation~\ref{eq:divergence-decomposition})
\begin{equation}\label{eq:proof-2-a}
    d_t(m^\ast\|m) = \sum_{m'} g(m';\fs{T}_{m'}).
\end{equation}
Furthermore, for every $m' \in \fs{M}$, one has that for all $\delta > 0$,
there is a $C \geq 0$, such that for all $t \in \nats$ and for all $\fs{T}
\subset \fs{N}_t$
\[
    \Bigl|
        g(m';\fs{T}) - G(m';\fs{T})
    \Bigr|
    \leq C(m)
\]
with probability $\geq 1-\delta'$. Applying this bound to the summands
in~(\ref{eq:proof-2-a}) yields the lower bound
\[
    \sum_{m'} g(m';\fs{T}_{m'})
    \geq \sum_{m'} \bigl( G(m';\fs{T}_{m'}) - C(m) \bigr)
\]
which holds with probability $\geq (1-\delta')^M$, where $M \define |\fs{M}|$.
Due to Inequality~\ref{eq:gibbs-ineq-sub-diver}, one has that for all $m' \neq
m^\ast$, $G(m';\fs{T}_{m'}) \geq 0$. Hence,
\[
    \sum_{m'} \bigl( G(m';\fs{T}_{m'}) - C(m) \bigr)
    \geq G(m^\ast;\fs{T}_{m^\ast}) - M C
\]
where $C \define \max_{m} \{C(m)\}$. The members of the set $\fs{T}_{m^\ast}$
are determined stochastically; more specifically, the $i^\text{th}$ member is
included into $\fs{T}_{m^\ast}$ with probability $P(m^\ast|\g{\hat{a}o}_{\leq
i})$. But since $m \notin [m^\ast]$, one has that $G(m^\ast;\fs{T}_{m^\ast})
\rightarrow \infty$ as $t \rightarrow \infty$ with probability $\geq 1-\delta'$
for arbitrarily chosen $\delta'>0$. This implies that
\[
    \lim_{t \rightarrow \infty} d_t(m^\ast\|m)
    \geq \lim_{t \rightarrow \infty} G(m^\ast;\fs{T}_{m^\ast}) - MC
    \nearrow \infty
\]
with probability $\geq 1 - \delta$, where $\delta>0$ is arbitrary and related
to $\delta'$ as $\delta = 1-(1-\delta')^{M+1}$. Using this result in the upper
bound for posterior probabilities yields the final result
\[
    0 \leq \lim_{t \rightarrow \infty} P(m|\g{\hat{a}o}_{\leq t})
    \leq \lim_{t \rightarrow \infty} \frac{P(m)}{P(m^\ast)}
        e^{-d_t(m^\ast\|m)}
    = 0.
\]
\end{proof}

\section{Consistency}

\begin{figure}[htbp]
\centering %
\begin{scriptsize}
\psfrag{h1}{$H_1$} \psfrag{p1}{$P_1$} %
\psfrag{h2}{$H_2$} \psfrag{p2}{$P_2$} %
\psfrag{ra1}{$A$} \psfrag{ra2}{$A$} %
\psfrag{rb1}{$B$} \psfrag{rb2}{$B$} %
\includegraphics[width=7cm]{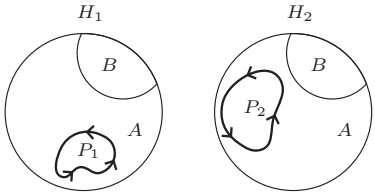}
\end{scriptsize}
\caption{An example of inconsistent policies. Both operation modes are in the
core of each other, but have different policies.}
\label{fig:consistency} %
\end{figure}

Even if an operation mode $m$ is in the core of $m^\ast$, i.e. given that $m$
is essentially indistinguishable from $m^\ast$ under $m^\ast$'s control, it can
still happen that $m^\ast$ and $m$ have different policies.
Figure~\ref{fig:consistency} shows an example of this. The hypotheses~$H_1$
and~$H_2$ share region~$A$ but differ in region~$B$. In addition, both
operation modes have their policies~$P_1$ and~$P_2$ respectively confined to
region~$A$. Note that both operation modes are in the core of each other.
However, their policies are different. This means that it is unclear whether
multiplexing the policies in time will ever disambiguate the two hypotheses.
This is undesirable, as it could impede the convergence to the right control
law.

Thus, it is clear that one needs to impose further restrictions on the mapping
of hypotheses into policies. With respect to Figure~\ref{fig:consistency}, one
can make the following observations:

\begin{enumerate}
  \item Both operation modes have policies that select subsets of region~$A$.
  Therefore, the dynamics in~$A$ are preferred over the dynamics in~$B$.
  \item Knowing that the dynamics in~$A$ are preferred over the dynamics in~$B$
  allows to drop region~$B$ from the analysis when choosing a policy.
  \item Since both hypotheses agree in region~$A$, \emph{they have to choose the same
  policy in order to be consistent in their selection criterion}.
\end{enumerate}

This motivates the following definition. An operation mode $m$ is said to be
\emph{consistent} with $m^\ast$ iff $m \in [m^\ast]$ implies that for all
$\varepsilon < 0$, there is a $t_0$, such that for all $t \geq t_0$ and all
$\g{ao}_{<t}a_t$,
\[
    \Bigl|
    P(a_t|m^\ast,\g{ao}_{\leq t}) - P(a_t|m^\ast,\g{ao}_{\leq t})
    \Bigr|
    < \varepsilon.
\]

In other words, if $m$ is in the core of $m^\ast$, then $m$'s policy has to
converge to $m^\ast$'s policy. Intuitively, this property parallels the
well-known \emph{sure-thing principle} of expected utility theory
\citep{Savage1954}. The following theorem shows that consistency is a
sufficient condition for convergence to the right control law.

\begin{theorem}
Let the set of operation modes of a controller be such that: for all $m \in
\fs{M}$ the divergence process $d_t(m^\ast\|m)$ is bounded; and for all $m, m'
\in \fs{M}$, $m$ is consistent with $m'$. Then,
\[
    P(a_t|\g{\hat{a}o}_{\leq t}) \rightarrow P(a_t|m^\ast,\g{ao}_{\leq t})
\]
almost surely as $t \rightarrow \infty$.
\end{theorem}

\begin{proof}
We will use the abbreviations $p_{m}(t) \define P(a_t|m, \g{\hat{a}o}_{\leq
t})$ and $w_m(t) \define P(m|\g{ao}_{\leq t})$. Decompose
$P(a_t|\g{\hat{a}o}_{\leq t})$ as
\begin{equation}\label{eq:proof-3-a}
    P(a_t|\g{\hat{a}o}_{\leq t})
    = \sum_{m \notin [m^\ast]} p_m(t) w_m(t)
    + \sum_{m \in [m^\ast]} p_m(t) w_m(t).
\end{equation}

The first sum on the right-hand side is lower-bounded by zero and upper-bounded
by
\[
    \sum_{m \notin [m^\ast]} p_m(t) w_m(t)
    \leq \sum_{m \notin [m^\ast]} w_m(t)
\]
because $p_m(t) \leq 1$. Due to Theorem~\ref{theo:vanishing-posterior}, $w_m(t)
\rightarrow 0$ as $t \rightarrow \infty$ almost surely. Given $\varepsilon'>0$
and $\delta'>0$, let $t_0(m)$ be the time such that for all $t \geq t_0(m)$,
$w_m(t) < \varepsilon'$. Choosing $t_0 \define \max_m \{t_0(m)\}$, the previous
inequality holds for all $m$ and $t \geq t_0$ simultaneously with probability
$\geq (1-\delta')^M$. Hence,
\begin{equation}\label{eq:proof-3-b}
    \sum_{m \notin [m^\ast]} p_m(t) w_m(t)
    \leq \sum_{m \notin [m^\ast]} w_m(t)
    < M \varepsilon'.
\end{equation}

To bound the second sum in~(\ref{eq:proof-3-a}) one proceeds as follows. For
every member $m \in [m^\ast]$, one has that $p_m(t) \rightarrow p_{m^\ast}(t)$
as $t \rightarrow \infty$. Hence, following a similar construction as above,
one can choose $t'_0$ such that for all $t \geq t'_0$ and $m \in [m^\ast]$, the
inequalities
\[
    \Bigl| p_m(t) - p_{m^\ast}(t) \Bigr| < \varepsilon'
\]
hold simultaneously for the precision $\varepsilon' > 0$. Applying this to the
first sum yields the bounds
\begin{align*}
    \sum_{m \in [m^\ast]} & \bigl( p_{m^\ast}(t) - \varepsilon' \bigr) w_m(t)
    \\ & \leq \sum_{m \in [m^\ast]} p_m(t) w_m(t)
    \\ & \leq \sum_{m \in [m^\ast]} \bigl( p_{m^\ast}(t) + \varepsilon' \bigr) w_m(t).
\end{align*}
Here $\bigl( p_{m^\ast}(t) \pm \varepsilon' \bigr)$ are multiplicative
constants that can be placed in front of the sum. Note that
\[
    1 \geq \sum_{m \in [m^\ast]} w_m(t)
    = 1 - \sum_{m \notin [m^\ast]} w_m(t)
    > 1 - \varepsilon.
\]
Diligently using of the above inequalities allows simplifying the lower and
upper bounds respectively:
\begin{equation}\label{eq:proof-3-c}
\begin{aligned}
    \bigl( p_{m^\ast}(t) - \varepsilon' \bigr)
    & \sum_{m \in [m^\ast]} w_m(t)
    > p_{m^\ast}(t) (1-\varepsilon') - \varepsilon'
    \\ & \geq p_{m^\ast}(t) - 2\varepsilon',
    \\ \bigl( p_{m^\ast}(t) + \varepsilon' \bigr)
    & \sum_{m \in [m^\ast]} w_m(t)
    \leq p_{m^\ast}(t) + \varepsilon'
    \\ & < p_{m^\ast}(t) + 2\varepsilon'.
\end{aligned}
\end{equation}

Combining the inequalities~(\ref{eq:proof-3-b}) and~(\ref{eq:proof-3-c})
in~(\ref{eq:proof-3-a}) yields the final result:
\[
    \Bigl| P(a_t|\g{\hat{a}o}_{\leq t}) - p_{m^\ast}(t) \Bigr| < 3\varepsilon' =
    \varepsilon,
\]
which holds with probability $\geq 1-\delta$ for arbitrary $\delta > 0$ related
to $\delta'$ as $\delta' = 1 - \sqrt[M]{1-\delta}$ and arbitrary
precision~$\varepsilon$.
\end{proof}

\section{Summary and Conclusions}\label{sec:conclusions}

The Bayesian control rule constitutes a promising approach to adaptive control
based on the minimization of the relative entropy of the causal I/O
distribution of a mixture controller from the true controller. In this work, a
proof of convergence of the Bayesian control rule to the true controller is
provided.

Analyzing the asymptotic behavior of a controller-plant dynamics could be
perceived as a difficult problem that involves the consideration of
domain-specific assumptions. Here it is shown that this is not the case: the
asymptotic analysis can be recast as the study of concurrent \emph{divergence
processes} that determine the evolution of the posterior probabilities over
operation modes, thus abstracting away from the details of the classes of I/O
distributions. In particular, if the set of operation modes is finite, then two
extra assumptions are sufficient to prove convergence. The first one,
\emph{boundedness}, imposes the stability of divergence processes under the
partial influence of the policies contained within the set of operation modes.
This condition can be regarded as an ergodicity assumption. The second one,
\emph{consistency}, requires that if a hypothesis makes the same predictions as
another hypothesis within its most relevant subset of dynamics, then both
hypotheses share the same policy. This relevance is formalized as the
\emph{core} of an operation mode.

The concepts and proof strategies developed in this work are appealing due to
their intuitive interpretation and formal simplicity. Most importantly, they
strengthen the intuition about potential pitfalls that arise in the context of
controller design. The approach presented in this work can also be considered
as a guide for possible extensions to infinite sets of operation modes. For
example, one can think of partitioning a continuous space of operation modes
into ``essentially different'' regions where representative operation modes
subsume their neighborhoods \citep{Grunwald2007}.

Finally, convergence proofs play a crucial r\^{o}le in the mathematical
justification of any new theory of control. Hopefully, this proof will
contribute to establish relative entropy control theories as solid alternative
formulations to the problem of adaptive control.

\vskip 0.2in
%GATHER{bibliography.bib}
\bibliographystyle{icml2010}
\small
\bibliography{bibliography}

\end{document}